\documentclass[runningheads]{llncs}
\usepackage{graphicx}

\newcommand{\eg}{e.g.,}
\newcommand{\ie}{i.e.,}
\newcommand{\ca}{\textit{CloudAcademy}}
\newcommand{\linkcode}{\url{https://github.com/lucabenedetto/text2props}}

\newcommand{\lowthres}{$\textrm{INF}$}
\newcommand{\upthres}{$\textrm{SUP}$}
\newcommand{\framework}{\textit{text2props}}

\begin{document}

\title{Introducing a framework to assess newly created questions with Natural Language Processing}

\titlerunning{Introducing a framework to assess newly created questions with NLP}

\author{
Luca Benedetto\inst{1} \orcidID{0000-0002-5113-4696} \and
Andrea Cappelli\inst{2} \and
Roberto Turrin\inst{2} \and
Paolo Cremonesi\inst{1} \orcidID{0000-0002-1253-8081}
}
\authorrunning{L. Benedetto et al.}

\institute{
Politecnico di Milano, Milan, Italy \\
\email{\{luca.benedetto,paolo.cremonesi\}@polimi.it} \and
Cloud Academy Sagl, Mendrisio, Switzerland \\
\email{\{andrea.cappelli,roberto.turrin\}@cloudacademy.com}
}
\maketitle
\begin{abstract}
Statistical models such as those derived from Item Response Theory (IRT) enable the assessment of students on a specific subject, which can be useful for several purposes (\eg{} learning path customization, drop-out prediction).
However, the questions have to be assessed as well and, although it is possible to estimate with IRT the characteristics of questions that have already been answered by several students, this technique cannot be used on newly generated questions.
In this paper, we propose a framework to train and evaluate models for estimating the difficulty and discrimination of newly created Multiple Choice Questions by extracting meaningful features from the text of the question and of the possible choices.
We implement one model using this framework and test it on a real-world dataset provided by \ca{}, showing that it outperforms previously proposed models, reducing by 6.7\% the RMSE for difficulty estimation and by 10.8\% the RMSE for discrimination estimation.
We also present the results of an ablation study performed to support our features choice and to show the effects of different characteristics of the questions' text on difficulty and discrimination.

\end{abstract}

\section{Introduction}\label{intro}
Modeling the skill level of students and how it evolves over time is known as Knowledge Tracing (KT), and it can be leveraged to improve the learning experience, for instance suggesting tailored learning content or detecting students in need of further support.
KT is most commonly performed with logistic models or neural networks.
Although neural models often reach the best accuracy in predicting the correctness of students' answers, they do not provide easy explanations of their predictions.
Logistic models such as Item Response Theory (IRT), instead, estimate latent traits of students and questions (\eg{} numerical values representing skill level and difficulty level) and use those to predict future answers.
IRT leverages the answers given by a student to a set of calibrated questions (\ie{} whose latent traits are known) to estimate her skill level, by finding the skill value that maximizes the likelihood of the observed results.
Questions' latent traits are non-observable parameters which have to be estimated and, if such estimation is not accurate, it affects the students' assessment and impacts the overall efficacy of the system (\eg{} suggesting wrongly targeted learning content).
Also, an accurate calibration of the questions allows to identify the ones that are not suited for scoring students because they cannot discriminate between different skill levels.
For instance, questions that are too difficult or too easy are answered in the same way by all the students, and questions that are unclear (\eg{} due to poor wording) are answered correctly or wrongly independently of the knowledge of the students.
Questions' latent traits are usually estimated with one of two techniques: they are either i) hand-picked by human experts or ii) estimated with pretesting.
Both approaches are far from optimal: manual labeling is intrinsically subjective, thus affected by high uncertainty and inconsistency; pretesting leads to a reliable and fairly consistent calibration but introduces a long delay before using new questions for scoring students \cite{yaneva2019predicting}.

Recent works tried to overcome the problem of calibrating newly-generated questions by proposing models capable of estimating their characteristics from the text: with this approach, it is possible to immediately obtain an estimation of questions' latent traits and, if necessary, this initial estimation can be later fine-tuned using students' answers.
However, most works targeted either the \textit{wrongness} or the \textit{p-value} of each question (\ie{} the fraction of wrong and correct answers, respectively), which are approximations of the actual difficulty; \cite{benedetto2020r2de} focus on latent traits as defined in IRT (\ie{} difficulty and discrimination).
This work introduces \framework{}, a framework to train and evaluate models for calibrating newly created Multiple-Choice Questions (MCQ) from the text of the questions and of the possible choices.
The framework is made of three modules for i) estimating ground truth latent traits, ii) extracting meaningful features from the text, and iii) estimating question's properties from such features.
The three modules can be implemented with different components, thus enabling the usage of different techniques at each step; it is also possible to use predefined ground truth latent traits, if available.
We show the details of a sample model implemented with \framework{} and present the results of experiments performed on a dataset provided by the e-learning provider \ca{}\footnote{\url{https://cloudacademy.com/}}.
Our experiments show an improvement in the estimation of both difficulty and discrimination: specifically, reaching a 6.7\% reduction in the RMSE for difficulty estimation (from 0.807 to 0.753) and 10.8\% reduction in the RMSE for discrimination estimation (from 0.414 to 0.369).
We also present an ablation study to empirically support our choice of features, and the results of an experiment on the prediction of students' answers, to validate the model using an observable ground truth.
The contributions of this work are:
i) the introduction of \framework{}, a framework to implement models for calibrating newly created MCQ,
ii) the implementation of a sample model that outperforms previously proposed models,
iii) an ablation study to support our choice of features in the sample model,
iv) publication of the framework's code to foster further research\footnote{\linkcode{}}.
This document is organized as follows: Section \ref{sec:related_work} presents the related works, Section \ref{sec:model} introduces \framework{}, Section \ref{sec:exp_setup} describes the dataset and the sample model, Section \ref{sec:results} presents the results of the experiments, Section \ref{sec:conclusions} concludes the paper.
\section{Related Work}\label{sec:related_work}

\subsection{Students' Assessment}
Knowledge Tracing (KT) was pioneered by Atkinson \cite{atkinson1972ingredients} and, as reported in a recent survey \cite{abyaa2019learner}, is most commonly performed with logistic models (\eg{} IRT \cite{wang2013bayesian}, Elo rating system \cite{verhagen2019toward}) or neural networks \cite{piech2015deep,abdelrahman2019knowledge}.
Recent works on students' performance prediction claim that Deep Knowledge Tracing (DKT) (\ie{} KT with neural networks \cite{piech2015deep}) outperforms logistic models in predicting the results of future exams \cite{zhang2017dynamic,chen2018prerequisite,abdelrahman2019knowledge,zhang2017incorporating}, but this advantage is not fully agreed across the community \cite{mao2018deep,dingdeep,yeung2018addressing,wilson2016back}.
Also, DKT predictions do not provide an explicit numerical estimation of the skill level of the students or the difficulty of the questions.
Recent works \cite{lee2019knowledge,yeung2019deep} attempted to make DKT explainable by integrating concepts analogous to the latent traits used in logistic models, but being much more expensive from a computational point of view and without reaching the same level of explainability as logistic models.
Thus, logistic models are usually chosen when interpretable latent traits are needed.
In this work, we use Item Response Theory (IRT) \cite{hambleton1991fundamentals}, that estimates the latent traits of students and questions involved in an exam.
We consider the two-parameters model, which associates to each item two scalars: the difficulty and the discrimination.
The difficulty represents the skill level required to have a 50\% probability of correctly answering the question, while the discrimination determines how rapidly the odds of correct answer increase or decrease with the skill level of the student.

\subsection{NLP for Latent Traits Estimation}
The idea of inferring properties of a question from its text is not new; however, most of previous works did not focus on difficulty estimation.
The first works focused on text readability estimation \cite{dubay2004principles,kintsch2014reading}.
In \cite{huang2019ekt} the authors use a neural network to extract from questions' text the topics that are assessed by each question.
Wang et al. in \cite{wang2014regularized} and Liu et al. in \cite{liu2013question} proposed models to estimate the difficulty of questions published in community question answering services leveraging the text of the question and some domain specific information which is not available in the educational domain, thus framing the problem differently.
Closer to our case are some works that use NLP to estimate the difficulty of assessment items, but most of them measured questions' difficulty as the fraction of students that answered incorrectly (\ie{} the \textit{wrongness}) or correctly (\ie{} the \textit{p-value}), which are arguably a more limited estimation than the IRT difficulty, as they do not account for different students' skill levels.
Huang et al. \cite{huang2017question} propose a neural model to predict the difficulty of ``reading'' problems in Standard Tests, in which the answer has to be found in a text provided to the students together with the question.
Their neural model uses as input both the text of the question and the text of the document, a major difference from our case.
Yaneva et al. in \cite{yaneva2019predicting} introduce a model to estimate the \textit{p-value} of MCQ from the text of the questions, using features coming from readability measures, word embeddings, and Information Retrieval (IR).
In \cite{qiu2019question} the authors propose a much more complex model, based on a deep neural network, to estimate the \textit{wrongness} of MCQ.
In \cite{benedetto2020r2de} the authors use IR features to estimate the IRT difficulty and the discrimination of MCQ from the text of the questions and of the possible choices.
All relevant related works experimented on private datasets and only \cite{benedetto2020r2de} focuses on IRT latent traits.
In this paper, we make a step forward with respect to previous research by introducing \framework{}, a modular framework to train and evaluate models for estimating the difficulty and the discrimination of MCQ from textual information.
Then, we implement a sample model with \framework{} and test is on a sub-sample of a private dataset provided by \ca{}.

\section{The framework}\label{sec:model}

\subsection{Data Format}\label{subsec:data_format}
The \textit{text2props} framework interacts with two datasets: i) the \textit{Questions} (\textit{Q}) dataset contains the textual information, ii) the \textit{Answers} (\textit{A}) dataset contains the results of the interactions between students and questions.
Specifically, \textit{Q} contains, for each question: i) ID of the question, ii) text of the MCQ, iii) text of all the possible choices, and iv) correctness of each choice; \textit{A}, instead, contains for each interaction: i) ID of the student, ii) ID of the question, iii) correctness of the answer, and iv) timestamp of the interaction.
The interactions in \textit{A} are used to obtain the ground truth latent traits of each question, which are used as target values while training the estimation of latent traits from textual information.

\subsection{Architecture}\label{subsec:modules}
Three modules compose \framework{}: i) an IRT estimation module to obtain ground truth latent traits, ii) a feature engineering module to extract features from text, and iii) a regression module to estimate the latent traits from such features.
At training time all the modules are trained, while only the feature engineering module and the regression module are involved in the inference phase.

\begin{figure}[ht]
\centering
\includegraphics[width=\textwidth]{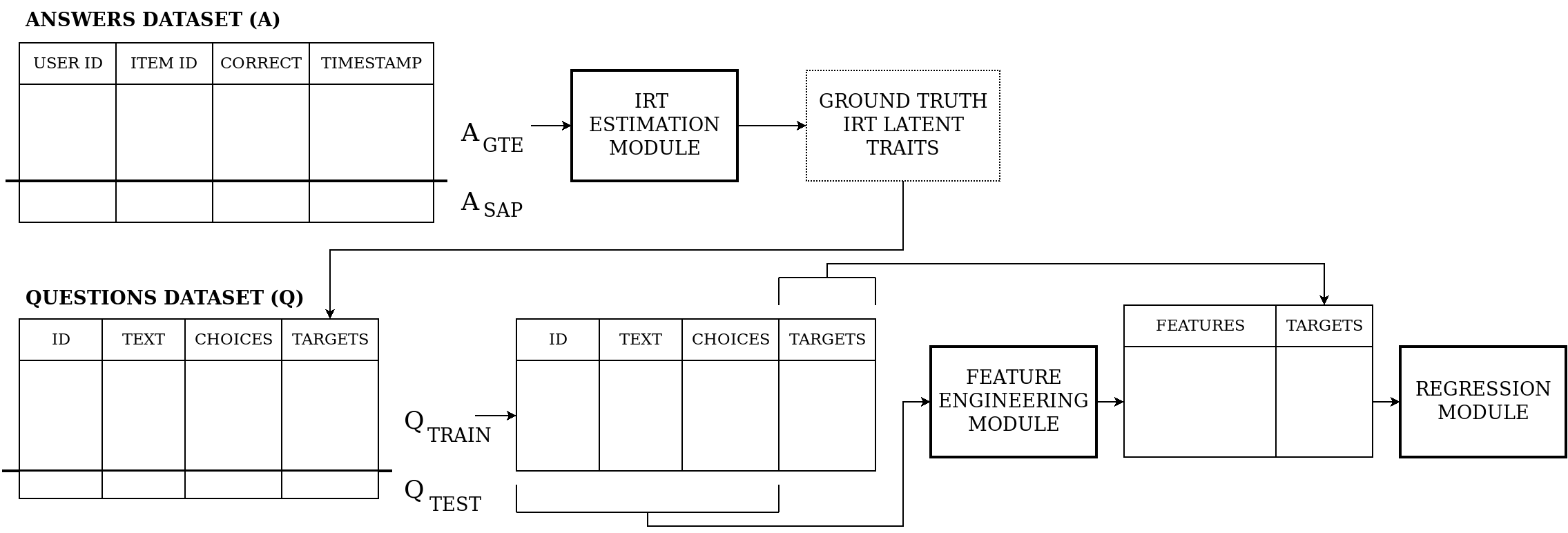}
\caption{Framework's architecture and interactions with the datasets during training.}\label{fig:framework_train}
\end{figure}
\begin{figure}[ht]
\centering
\includegraphics[width=.8\textwidth]{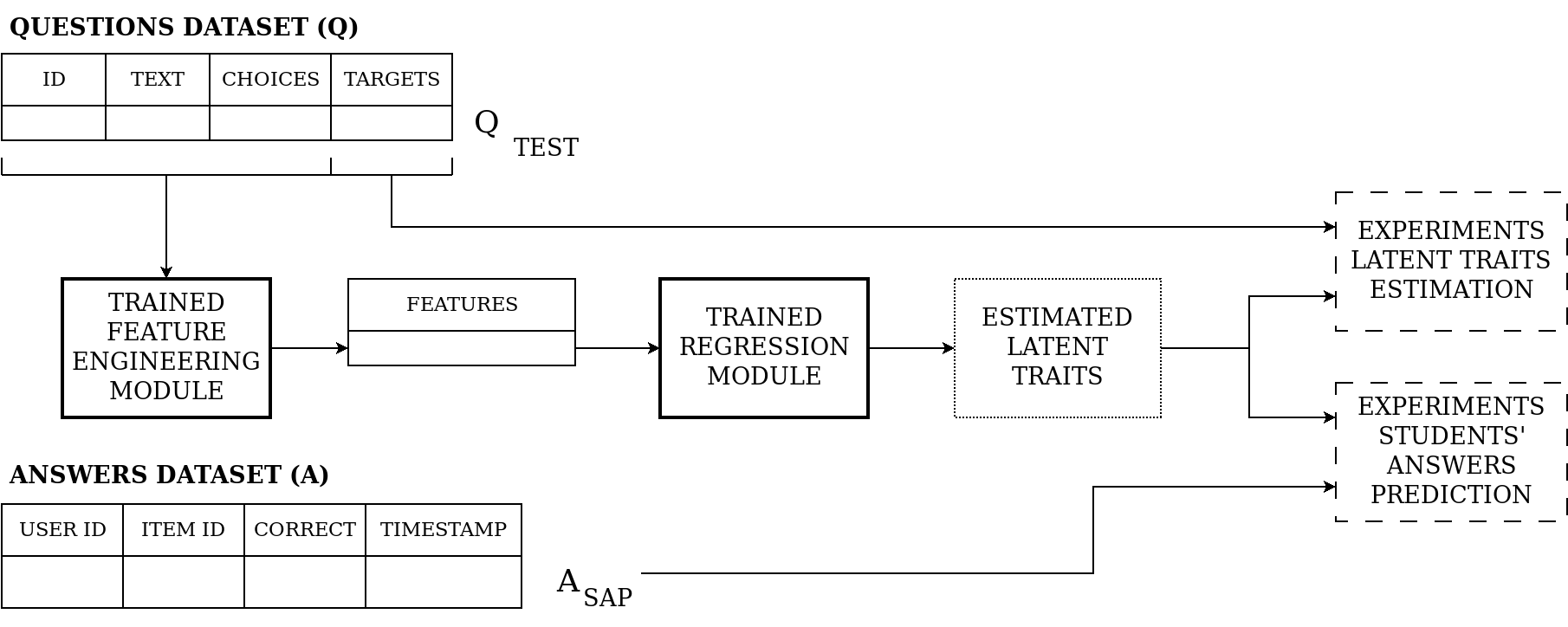}
\caption{Framework's architecture and interactions with the datasets during inference.}\label{fig:framework_test}
\end{figure}
Figure \ref{fig:framework_train} shows how the three modules interact with the datasets during \textbf{training}.
A split stratified on the questions is performed on \textit{A}, producing the dataset for estimating the ground truth latent traits (A\textsubscript{GTE}) and the dataset for evaluating students' answers prediction (A\textsubscript{SAP}).
This is done in order to have all the questions in both datasets and, therefore, be able to obtain the ground truth latent traits of all the questions from A\textsubscript{GTE} and, later, perform the experiments on students' answers prediction using previously unseen interactions.
The ground truth latent traits obtained with the IRT estimation module from A\textsubscript{GTE} are then stored in \textit{Q}, in order to be used as target values in the regression module.
Then, a split is performed on \textit{Q}, obtaining the dataset used to train the feature engineering and regression modules (Q\textsubscript{TRAIN}) and the dataset to test them (Q\textsubscript{TEST}).
Lastly, the textual information of Q\textsubscript{TRAIN} is used by the feature engineering module to extract numerical features, which are then used together with the ground truth latent traits to train the regression module.

During the \textbf{inference} phase, pictured in Figure \ref{fig:framework_test}, the trained feature engineering module is fed with the textual information of the questions in Q\textsubscript{TEST}, and extracts the features that are given to the trained regression module to estimate the latent traits.
These estimated latent traits can then be used for evaluating i) latent traits estimation, directly comparing them with the ground truth latent traits (in Q\textsubscript{TEST}), and ii) students' answers prediction, comparing the predictions with the true answers (in A\textsubscript{SAP}).


\section{Experimental Setup}\label{sec:exp_setup}

\subsection{Sample Model}\label{subsec:implem}
In order to implement a model using \framework{}, it is sufficient to define the three modules.
In the sample model used for the experiments, the calibration module performs the estimation of the IRT difficulty and discrimination of each question; these two latent traits are then used as ground truth while training the part of the model that performs the estimation from text.
The regression module contains two Random Forests to estimate the difficulty and discrimination.
The feature engineering module is made of three components to compute: i) readability features, ii) linguistic features, iii) Information Retrieval features.

\begin{itemize}
\item \textbf{Readability indexes} are measures designed to evaluate how easy a text is to understand, thus they can prove useful for estimating question's properties, as suggested in \cite{yaneva2019predicting}.
In particular, we use: \textit{Flesch Reading Ease} \cite{flesch1948new}, \textit{Flesch-Kincaid Grade Level} \cite{kincaid1975derivation}, \textit{Automated Readability Index} \cite{senter1967automated}, \textit{Gunning FOG Index} \cite{gunning1968technique}, \textit{Coleman-Liau Index} \cite{coleman1965understanding}, and \textit{SMOG Index} \cite{mc1969smog}.
All these indexes are computed with deterministic formulas from measures such as the number of words and the average word length.

\item The usage of \textbf{linguistic features} is motivated by \cite{dubay2004principles}, in which they proved useful for readability estimation.
The following features are used: \textit{Word Count Question}, \textit{Word Count Correct Choice}, \textit{Word Count Wrong Choice}, \textit{Sentence Count Question}, \textit{Sentence Count Correct Choice}, \textit{Sentence Count Wrong Choice}, \textit{Average Word Length Question}, \textit{Question Length divided by Correct Choice Length}, \textit{Question Length divided by Wrong Choice Length}.

\item The choice of \textbf{Information Retrieval} (IR) features is supported by previous research \cite{benedetto2020r2de} and by the idea that there must be a relation between the latent traits of a MCQ and the words that appear in the text.
We i) preprocess the texts using standard steps of NLP \cite{manning1999foundations}, ii) consider both the text of the question and the text of the possible choices by concatenating them, and iii) use features based on Term Frequency-Inverse Document Frequency (TF-IDF).
However, instead of keeping only the words whose frequency is above a certain threshold (as in \cite{benedetto2020r2de}), we define two thresholds - tuned with cross-validation - to remove i) corpus-specific stop words (\ie{} words with frequency above \upthres{}) and ii) very uncommon words (\ie{} with frequency below \lowthres{}).
\end{itemize}

\subsection{Experimental Dataset}\label{subsec:dataset}
All previous works experimented on private data collections \cite{yaneva2019predicting,benedetto2020r2de,huang2017question,qiu2019question} and, similarly, we evaluate this framework on a private dataset, which is a sub-sample of real world data coming from the e-learning provider \ca{}.
Dataset \textit{Q} contains about 11K multiple-choice questions and they have 4 possible answers; for some questions, there is more than one correct answer and, in that case, the student is asked to select all the correct choices.
Dataset \textit{A}, which is used for estimating the ground truth latent traits and for the experiments on students' answers prediction, contains about 2M answers.
Also, it is filtered in order to keep only the students and the questions that appear in at least 100 different interactions; thus we assume that the IRT-estimated latent traits are accurate enough to be used as ground truth for this study.

\section{Results}\label{sec:results}

\subsection{Latent Traits Estimation}\label{subsec:lte}
The sample model used for the comparison with the state of the art was chosen from a pool of models, all implemented with  \framework{}.
All these models had the same IRT estimator module and the same feature engineering module, containing the three components described in Section \ref{subsec:implem}, but they were implemented with different algorithms in the regression module: specifically, we tested Random Forests (RF), Decision Trees (DT), Support Vector Regression (SVR), and Linear Regression (LR).
For each model, hyperparameter tuning was performed via 10-fold randomized cross-validation \cite{bergstra2012random}.
The results of this preliminary experiments for choosing the sample model are displayed in Table \ref{tab:res_diff}, presenting for each candidate model the Root Mean Square Error (RMSE) and the Mean Absolute Error (MAE) for difficulty estimation and discrimination estimation, separately on a validation set held-out from the test set and on the remaining test set.
The two errors measure how accurate the sample model is by comparing the latent traits (\ie{} difficulty and discrimination) estimated from text with the ground truth values obtained with IRT estimation. 
\begin{table}
\centering
\caption{Preliminary experiments for choosing the sample model.}\label{tab:res_diff}
\begin{tabular}{c c c c c c c c c}
\hline
 & \multicolumn{4}{c}{Difficulty estimation} & \multicolumn{4}{c}{Discrimination estimation} \\
 & \multicolumn{2}{c}{Validation set} & \multicolumn{2}{c}{Test set} & \multicolumn{2}{c}{Validation set} & \multicolumn{2}{c}{Test set}\\
Regression module & RMSE & MAE & RMSE & MAE                                     & RMSE & MAE & RMSE & MAE \\
\hline
RF       & 0.739 & 0.575 & \textbf{0.753} & \textbf{0.587}           & 0.393 & 0.296 & \textbf{0.369} & \textbf{0.287} \\
DT       & 0.748 & 0.586 & 0.826          & 0.636                    & 0.393 & 0.295 & 0.375 & 0.290\\
SVR      & 0.797 & 0.632 & 0.804          & 0.629                    & 0.394 & 0.298 & 0.379 & 0.296\\
LR       & 0.752 & 0.599 & 0.779          & 0.607                    & 0.397 & 0.298 & 0.378 & 0.293\\
\hline
Majority & -     & -     & 0.820          & 0.650                    & -     & -     & 0.502 & 0.427\\
\hline
\end{tabular}
\end{table}
As baseline, we consider a majority prediction, which assigns to all the questions the same difficulty and discrimination, obtained by averaging the training latent traits.
All the models outperform the majority baseline, and the RF leads to the best performance in both cases; thus, that is the model which will be used as sample model for the rest of the experiments and the comparison with the state of the art.

Table \ref{tab:sota_lte} compares the model implemented with \framework{} with the state of the art for difficulty and discrimination estimation.
Considering difficulty estimation, our model reduces the RMSE by 6.7\% (from 0.807 to 0.753) with respect to R2DE, which was implemented using the code publicly available\footnote{\url{https://github.com/lucabenedetto/r2de-nlp-to-estimating-irt-parameters}}, re-trained and tested on the new dataset.
\begin{table}[ht]
\centering
\caption{Comparison with state of the art.}\label{tab:sota_lte}
\begin{tabular}{ c c c c c c c c c}
\hline
                                  & & \multicolumn{3}{c}{Difficulty estimation} & & \multicolumn{3}{c}{Discrimination estimation} \\
Model                                     & & Range & RMSE & Relative RMSE              & & Range & RMSE & Relative RMSE \\
\hline
Our model                                 & & [-5; 5]  & \textbf{0.753}  & \textbf{7.53\%}       & & [-1; 2.5] & \textbf{0.369} & \textbf{9.22\%} \\
R2DE \cite{benedetto2020r2de}             & & [-5; 5]  & 0.807  & 8.07\%                & & [-1; 2.5] & 0.414 & 10.35\% \\
\hline
Qiu et al. \cite{qiu2019question}         & & [0; 1]   & 0.1521 & 15.21\%               & & - & - & - \\
Huang et al. \cite{huang2017question}     & & [0; 1]   & 0.21   & 21\%                  & & - & - & - \\
Yaneva et al. \cite{yaneva2019predicting} & & [0; 100] & 22.45  & 22.45\%               & & - & - & - \\
\hline
\end{tabular}
\end{table}
The other works experimented on private datasets and could not be directly re-implemented on our dataset, therefore a comparison on the same dataset was not straightforward; however, as suggested in \cite{benedetto2020r2de}, we can still gain some insight by performing a comparison on the Relative RMSE, which is defined as:
$ \textrm{RMSE} / (\textrm{difficulty}_{\textrm{max}} - \textrm{difficulty}_{\textrm{min}})$.
The Relative RMSE of the sample model is smaller than the ones obtained in previous research and, although this does not guarantee that it would perform better than the others on every dataset, it suggests that it might perform well.
The part of the table about discrimination estimation contains only two lines since this and R2DE are the only works that estimate both the difficulty and the discrimination.
Again, our model outperforms R2DE, reducing the RMSE from 0.414 to 0.369.

\subsection{Students' answers prediction}\label{subsec:pp}
The accuracy of latent traits estimation is commonly evaluated by measuring the error with respect to ground truth latent traits estimated with IRT. 
However, although IRT is a well-established technique, such latent traits are non observable properties, and we want to validate our model on an observable ground truth as well, therefore we evaluate the effects that it has in predicting the correctness of students' answers.
Students' Answers Prediction (SAP) provides an insight on the accuracy of latent traits estimation because questions' latent traits are a key element in predicting the correctness of future answers.
Indeed, given a student $i$ with estimated skill level $\tilde{\theta}_i$ and a question $j$ with difficulty $b_j$ and discrimination $a_j$, the probability of correct answer is computed as 
\begin{equation}
P_C = \frac{1}{1 + e^{-1.7a_j \cdot (\tilde{\theta}_i - b_j)}}
\end{equation}
The skill level $\tilde{\theta}_i$ is estimated from the answers previously given by the student:
\begin{equation}
\tilde{\theta}_i = \max_{\theta} \left[ \prod_{q_j \in Q_{C}}\frac{1}{1 + e^{-1.7a_j \cdot (\theta - b_j)}} \cdot \prod_{q_j \in Q_{W}} \left( 1 - \frac{1}{1 + e^{-1.7a_j \cdot (\theta - b_j)}}\right) \right]
\end{equation}
where $Q_C$ and $Q_W$ are sets containing the exam's questions correctly and wrongly answered, respectively.

Known the ordered sequence of interactions, SAP is performed as follows:
\begin{enumerate}
\item given the latent traits of a question ($b_j$, $a_j$) and the student's estimated skill level ($\tilde{\theta}_i$, possibly unknown), the probability of correct answer is computed;
\item if the probability is greater than 0.5 we predict a correct answer;
\item the real answer is observed and compared to the prediction (this is the comparison used to compute the evaluation metrics); 
\item the real answer is used to update the estimation of the student's skill level; 
\item these steps are repeated for all the items the student interacted with.
\end{enumerate}

By using in the two equations above latent traits coming from different sources, we compare the accuracy of SAP obtained i) with the latent traits estimated with our model, and ii) ground truth IRT latent traits.
Table \ref{tab:perf_pred} displays the results of the experiment, showing also as baseline a simple majority prediction.
\begin{table}
\centering
\caption{Students' asnwers prediction.}\label{tab:perf_pred}
\begin{tabular}{ c c c c c c c }
\hline
 & & & \multicolumn{2}{c}{Correct} & \multicolumn{2}{c}{Wrong}\\
Model & AUC & Accuracy & Precision & Recall & Precision & Recall \\
\hline
IRT & 0.74 & 0.683 & 0.744 & 0.735 & 0.589 & 0.599 \\
Our model & 0.66 & 0.630 & 0.707 & 0.678 & 0.521 & 0.555 \\
Majority & 0.50 & 0.613 & 0.613 & 1.0 & - & 0.000 \\
\hline
\end{tabular}
\end{table}
As metrics, we use Area Under Curve (AUC), accuracy, precision and recall on correct answers, and precision and recall on wrong answers.
The table shows that our model performs consistently better than the majority baseline and fairly closely to IRT - which is a upper threshold - suggesting that the estimation of latent traits from text can be successfully used as initial calibration of newly generated items.
However, it might still be convenient to fine-tune such estimation when the data coming from student interactions becomes available.

\subsection{Ablation Study}\label{subsec:ablation_study}
The objective of this ablation study is to i) empirically support our choice of features and ii) assess the impact of specific features on the estimation.
Table \ref{tab:ablation} presents the RMSE and the MAE for difficulty estimation and discrimination estimation.
In all cases, we use Random Forests in the regression module, since it seemed to be the most accurate and robust approach, according to the preliminary experiments; as baseline, we consider the majority prediction.
\begin{table}
\centering
\caption{Ablation study.}\label{tab:ablation}
\begin{tabular}{c c c c c c c c c c c}
\hline
 & & \multicolumn{4}{c}{Difficulty Estimation} & & \multicolumn{4}{c}{Discrimination Estimation}\\
Features & & \lowthres{} & \upthres{} & RMSE & MAE                       & & \lowthres{} & \upthres{} & RMSE & MAE\\
\hline
IR + Ling. + Read.  & & 0.02 & 0.92 &  \textbf{0.753} & \textbf{0.587}   & &  0.02 & 0.96 &  \textbf{0.369} & \textbf{0.287}\\
IR + Ling.          & & 0.02 & 0.90 & 0.754 & \textbf{0.587}             & &  0.02 & 0.98 & 0.370 & \textbf{0.287} \\
IR + Read.          & &  0.02 & 0.94 & 0.766 & 0.597                     & &  0.02 & 0.98 & 0.370 & 0.288\\
IR                  & &  0.00 & 0.92 & 0.758 & \textbf{0.587}            & &  0.02 & 0.96 &  0.372 & 0.289\\
Read + Ling         & &  - & - & 0.791 & 0.618                           & &  - & - & 0.373 & 0.291 \\
Readability         & &  - & - & 0.794 & 0.619                           & &  - & - & 0.374 & 0.292\\
Linguistic          & &  - & - & 0.791 & 0.620                           & &  - & - & 0.375 & 0.292 \\
Majority            & & - & - & 0.820 & 0.650                            & &  - & - & 0.502 & 0.427\\
\hline
\end{tabular}
\end{table}
The combination of all the features leads to the smallest errors, thus suggesting that all the features bring useful information.
The IR features seem to provide the most information when considered alone: this is reasonable, since they have two parameters that can be tuned to improve the performance.
The smallest error is usually obtained when some terms are removed from the input text; most likely, both corpus specific stop-words and terms which are too rare only introduce noise.
It is interesting to notice that readability and linguistic features seem to be more useful for discrimination than difficulty estimation since, when used alone, they perform similarly to the best performing features.

\section{Conclusions}\label{sec:conclusions}
In this paper we introduced \framework{}, a framework that allows the training and evaluation of models for calibrating newly created Multiple-Choice Questions from textual information.
We evaluated a sample model implemented with \framework{} on the tasks of latent traits estimation and students' answers prediction, showing that models implemented with this framework are capable of providing an accurate estimation of the latent traits, thus offering an initial calibration of newly generated questions, which can be fine-tuned when student interactions become available.
Our model outperformed the baselines reaching a 6.7\% reduction in the RMSE for difficulty estimation and 10.8\% reduction in the RMSE for discrimination estimation.
As for students' answers prediction, it improved the AUC by 0.16 over the majority baseline, and performed fairly close to the prediction made with IRT latent traits (which is an upper threshold), having an AUC 0.08 lower.
Lastly, the ablation study showed that all features are useful for improving the estimation of the latent traits from text, as the best results are obtained when combining all of them.
Future works will focus on exploring the effects of other features on the estimation of latent traits (\eg{} word embeddings, latent semantic analysis) and testing the capabilities of this framework to estimate other question's properties.
Also, future work should focus on the main limitation of \framework{}, consisting in the fact that it forces the implemented models to have the three-modules architecture presented here; in this case the model implemented with this framework proved effective, but it is not guaranteed that it would work similarly well in other situations.

\bibliographystyle{splncs04}
\bibliography{bib}

\end{document}